%% file: root.tex
\documentclass[letterpaper, 10 pt, conference]{ieeeconf}  

\IEEEoverridecommandlockouts                              

\usepackage{graphicx}
\usepackage{amsmath}
\usepackage{amssymb}
\usepackage{booktabs}
\usepackage{paralist, tabularx}
\usepackage{xcolor}

\title{\LARGE \bf
MENTOR: Multilingual tExt detectioN TOward leaRning by analogy
}

\author{Hsin-Ju Lin, Tsu-Chun Chung, Ching-Chun Hsiao, Pin-Yu Chen, Wei-Chen Chiu, and Ching-Chun Huang
\thanks{All authors except Dr. Pin-Yu Chen (affiliated to IBM research) are with the Department of Computer Science and Engineering, National Yang Ming Chiao Tung University, Hsinchu, Taiwan.}%
}

\begin{document}

\maketitle
\thispagestyle{empty}
\pagestyle{empty}

\newcommand{\CCH}[1]{{\color{red}#1}\normalfont}
\begin{abstract}
Text detection is frequently used in vision-based mobile robots when they need to interpret texts in their surroundings to perform a given task. For instance, delivery robots in multilingual cities need to be capable of doing multilingual text detection so that the robots can read traffic signs and road markings. Moreover, the target languages change from region to region, implying the need of efficiently re-training the models to recognize the novel/new languages. However, collecting and labeling training data for novel languages are cumbersome, and the efforts to re-train an existing/trained text detector are considerable. 
Even worse, such a routine would repeat whenever a novel language appears. This motivates us to propose a new problem setting for tackling the aforementioned challenges in a more efficient way: ``We ask for a generalizable multilingual text detection framework to detect and identify both seen and unseen language regions inside scene images without the requirement of collecting supervised training data for unseen languages as well as model re-training''. To this end, we propose ``MENTOR'', the first work to realize a learning strategy between zero-shot learning and few-shot learning for multilingual scene text detection. During the training phase, we leverage the ``zero-cost'' synthesized printed texts and the available training/seen languages to learn the meta-mapping from printed texts to language-specific kernel weights. Meanwhile, dynamic convolution networks guided by the language-specific kernel are trained to realize a detection-by-feature-matching scheme. In the inference phase, ``zero-cost'' printed texts are synthesized given a new target language. By utilizing the learned meta-mapping and the matching network, our ``MENTOR'' can freely identify the text regions of the new language.
Experiments show our model can achieve comparable results with supervised methods for seen languages and outperform other methods in detecting unseen languages.
\end{abstract}

\input{1_introduction} 
\input{2_related_work}
\input{3_method}

\input{4_experiment}

\input{5_conclusion}

\addtolength{\textheight}{-12cm}   









{\small
\bibliographystyle{ieee_fullname}
\bibliography{egbib}
}

\end{document}

%% file: 1_introduction.tex
\section{INTRODUCTION}

As cross-border travel and the popularity of social networks are increasing day by day, many languages, such as Chinese, French, Russian, Spanish, Arabic, and English, are listed as the world's lingua franca. Thereby, the street view texts in many countries are no longer limited to English or domestic language but also non-official ones. Such changes make multilingual scene text detection an essential issue for autonomous robot or vehicle navigation.

Previous multilingual scene text detection methods usually inherit the pre-trained model designed for detecting English texts with modifications. As a result, the performance of these methods could be limited due to the diverse characteristics of different languages. For example, Chinese, Japanese, and Korean are often written vertically, with large spacing between characters, disparate aspect ratios, and square shapes. Thus, a rich and well-labeled multiple-language text dataset becomes emergent to ensure a trained model can capture various properties of multilingual texts.

\begin{figure}[t]
  \centering
  \includegraphics[width=1\linewidth]{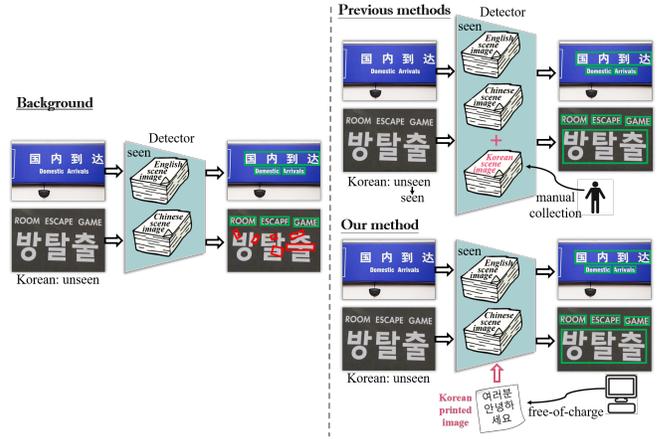}
  \vspace{-2em}
  \caption{\textbf{Left}: The detection results of an existing detector for multilingual scene text could be problematic when the model encounters a language that hasn’t been learned previously. \textbf{Upper-right}: For understanding unseen languages, previous approaches often require large-scale datasets collected and annotated by humans to retain the models for recognizing new languages. \textbf{Lower-right}: Free-of-charge printed text images are the only data required for our method to detect texts of unseen languages without any effort on model re-training.}
  \label{fig:fig1} 
  \vspace{-1em}
\end{figure}

However, a universal text dataset containing all languages for supervised training does not exist. Furthermore, as the most popular language in the world, English texts are usually the majority in current public multilingual scene datasets. Without careful training, the trained model would tend to emphasize a specific language and overlook others (i.e. biased), leading to a performance drop in multilingual scenarios.

On the other hand, a practical multilingual text detecter is expected to be extendable and can be applied to new languages easily and quickly. Model finetuning using labeled samples of new languages is a common approach to address the issue. However, labeled data collection is expensive, and model finetuning is also likely to cause catastrophic forgetting on the learned languages. 
In summary, the current challenges are as follows: 1) A universal text dataset containing all languages for supervised learning is not available, and it would take a heavy workload to create one; 2) The imbalanced text training dataset raises training difficulty; 3) The detection model must be re-trained to detect new languages. These challenges eventually inspire us to define a new problem setting: ``For generalized multilingual text detection, the model should identify both unseen and unseen language regions in the scene images, without relying on the training data of unseen languages and model retraining.''

The conceptual difference between our new setting and the conventional one is illustrated in Figure~\ref{fig:fig1}. To tackle the new problem setting, few-shot learning (FSL)~\cite{ltc,proto,sia,relation} seems to a possible solution which can walk around the challenge of model retraining but still requires supervised dataset collection for novel languages (even few-shots).
Another potential technique to address the problem is zero-shot learning (ZSL)~\cite{iap}, where some approaches indeed do not require any dataset collection for the new recognition targets nor any model retraining. 
Nevertheless, such free lunch comes from an important prerequisite for adopting the auxiliary knowledge (e.g. a set of predefined attributes shared among different object classes) as a bridge to connect among multilingual texts, in which it is not trivial to obtain (and does not exist to the best our knowledge) thus making ZSL unsuitable for our new problem setting.

To this end, we propose ``MENTOR: Multilingual tExt detectioN TOward leaRning by analogy'', that realizes a novel learning strategy in-between zero-shot learning and few-shot learning. To be detailed, during the training phase, based on the idea that the character set of a new language is usually the well-known prior knowledge no matter whether the scene text images of such language is seen or not, we leverage the ``zero-cost'' synthesized printed texts and the available seen languages to learn the meta-mapping function from printed texts to language-specific kernel weights through a latent representation. Next, a dynamic convolution network guided by the language-specific kernel is designed and trained to realize target language text detection in a  detection-by-matching manner. Here, the learned meta-mapping plays two roles: 1) It can implicitly describe the synthesized printed texts from a given language as a language-specific latent representation; 2) It further converts the latent representation to a language-specific kernel for feature matching.  
In the inference phase,``zero-cost'' printed texts are synthesized for a seen/unseen language. By utilizing the learned meta-mapping and the matching network, our ``MENTOR'' can freely identify the text regions of both seen and unseen languages in testing images, where the only materials we need for adapting to new languages are merely ``zero-cost'' printed texts. Thus, our method works like FSL (from high-level perspective) but requires no retraining and reliefs model forgetting. Our main contributions are summarized as:
\begin{itemize}
\item We proposed a new problem setting for generalized multi-language scene text detection.
\item Our multilingual detector, MENTOR, is able to generalize the matching relationship between external information (i.e., extracted from printed texts) and the features of scene text to detect unseen languages.
\item We design a data augmentation method to balance the multilingual training dataset. Also, we develope an efficient way to generate images of printed texts as the external information of languages.
\item Experiments show that our model performs comparably with supervised multilingual detection models.
\end{itemize}

%% file: 2_related_work.tex
\begin{figure*}[t]
  \centering
  \includegraphics[width=1\linewidth]{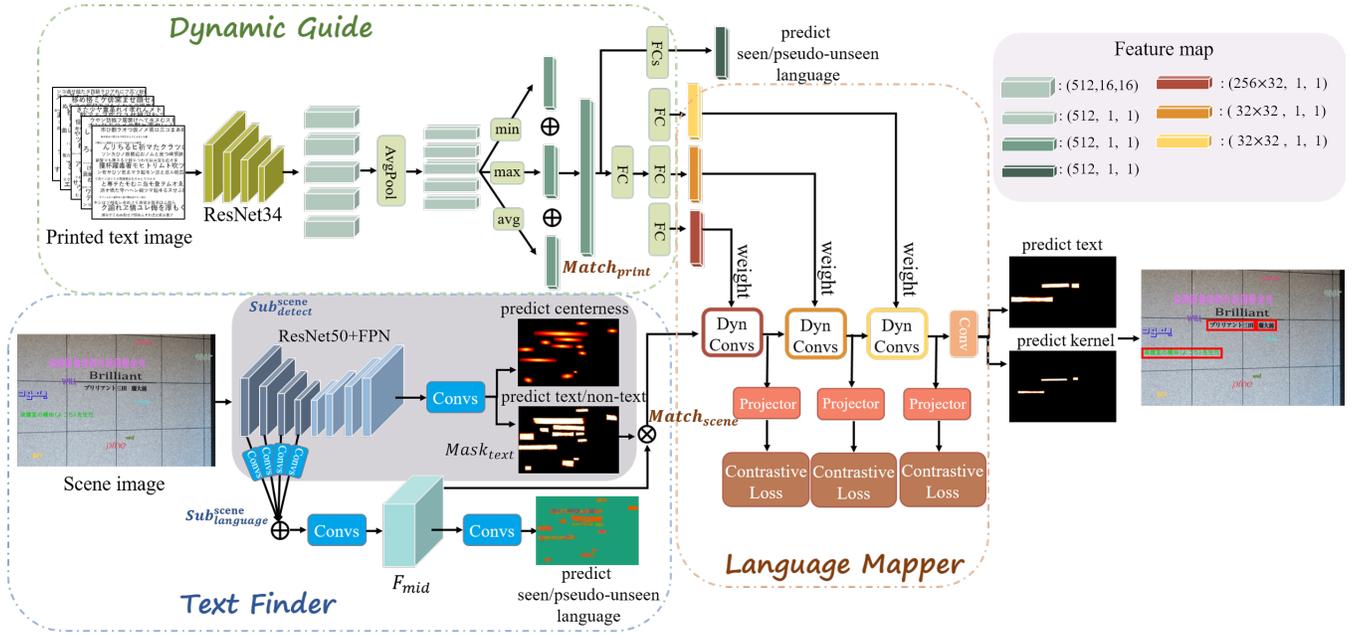}

  \caption{Overview of our proposed framework ``\textbf{MENTOR}''. Assuming that the target language is Japanese, we will prepare 5 sheets with printed Japanese texts (which are almost zero-cost) as the input of Dynamic Guide~(DG) while the scene text image to be examined is the input of our Text Finder~(TF). DG extracts the Japanese text attributes as auxiliary information to the prediction. TF is split into two paths after the backbone extraction. The Gray block is a language-agnostic detection sub-branch that is trained separately and outputs the centerness map and the text map, $Mask_{text}$. The lower sub-branch of TF predicts scene text language, and the middle feature $F_{mid}$ will multiply with $Mask_{text}$ to send the text-only areas to the Language Mapper~(LM). In the Language Mapper, the transferred Japanese text attributes are used as the weights of different dynamic kernels, and convoluted with the scene text features. Through progressive comparison, the regions where the scene text and the printed text have similar language characteristics are selected. Later, the Japanese detection result is obtained. The detection flow can be directly applied to detect other unseen languages without retraining the model.}
  \label{fig:framework} 
  \vspace{-1em}
\end{figure*}

\section{RELATED WORKS}
    Scene text detection in a complicated environment is an essential and practical task, and becoming a critical preprocessing step for robot navigation and many intelligent services (e.g. scene text recognition/translation and environmental understanding). Here we categorize the related works into two fields: monolingual and multilingual text detection.

\noindent\textbf{Monolingual Scene Text Detection.}~
    Although scene text detection is not a new task, most related works focus on developing advanced deep-learning networks to improve the accuracy of monolingual scene text detection. Among them, scene text detection for English gains more attention and achieves great success due to richer supervised datasets. In the early stage, scene text detection and object detection were treated as similar tasks. Thus, region proposal networks designed for object detection \cite{rcnn,pyramid,anchor} are directly utilized for text detection. However, text regions have flexible and variant shapes that differ from rigid object detection, raising a new challenge for using object detection approaches for text localization. 
    
    The follow-up works hence propose to use regression-based methods (estimating the corners of the bounding box for a text region) for avoiding improper prior constraints on the object aspect ratio which are typically adopted for object detection. 
    While \cite{fots,geometry} advance to take into account the geometric property of scene text regions to predict the rotation angles of bounding boxes, most regression-based methods still limit the text shape to be rectangular.

    To detect more accurate text regions for text recognition, researchers view the detection of irregular text as a new challenge. Current methods \cite{progress,bezier,craft,accurate} no longer ask text boxes to be quadrilateral. For example, the work \cite{bezier} generates a text box by predicting the parameter of a Bezier-Curve. And \cite{progress} proposes to identify a rough text box via estimating controllable boundary points, followed by a refinement step to output more accurate and irregular outline.

    On the other hand, segmentation-based methods are proposed to solve the problem of boundary ambiguity caused by adjacent text regions. Instead of directly determining the text regions, \cite{pse,mask} base on text segmentation to specify shrunk text regions for separating text instances. Then, the original text regions are restored by dilating the shrunk text using the Vatti clipping algorithm \cite{vatti}. 
    
\noindent\textbf{Multilingual Scene Text Detection.}~
    Recently, multilingual scene text detection begins to receive attention. Some works expand the aforementioned monolingual approaches and training strategies to detect multilingual texts directly \cite{position,explore}. However, the performance is typically unsatisfactory owing to the variant text properties among multiple languages. For instance, the scene texts in Chinese have splitting long lines, which are quite different from other languages. Accordingly, methods specifically designed for multilingual scene text detection are proposed. The authors in \cite{context} propose context attention that includes global and local contexts to explore better contextual information for different languages, while \cite{look} introduces a novel iterative module to solve extremely long texts by continuous refinement. In contrast, \cite{most} developes a text feature alignment module, which dynamically adjusts receptive fields to tackle the problem of different aspect ratios for various languages.
    
    However, these multilingual text detection methods are typically trained supervisedly and focus only on detecting texts in the pre-defined language set. Given a testing image contains texts of unseen languages, the conventional methods would fail the detection task. Thus, training a generalized detection model that can identify both seen and unseen languages is still an open issue.

%% file: 3_method.tex
\section{Methodology}
    Our goal is to design a detector following the new setting, which can detect any specified seen or unseen languages from the multilingual scene text image without requiring the network to learn all languages in advance. The proposed ``MENTOR'' framework is illustrated in Figure  \ref{fig:framework}, which is composed of three main components: Dynamic Guide (DG), Text Finder (TF), and Language Mapper (LM). The DG, TF, and LM modules work together in our proposed design to achieve our objective. The DG module is responsible for generating representational properties of different languages from printed text images. The TF module is indispensable in discovering the candidate text regions from the scene image. Finally, the LM module matches the printed text attributes (extracted by DG) with the scene text features (extracted by TF) to check for any similarity between the text region in the scene image and the printed text features.  
 
    In the following subsections, we first explain how to generate the ``zero-cost'' external information (i.e., images of printed texts) and how to synthesize balanced text instances to augment scene images. Next, we introduce three main components in detail. Finally, the design of our objectives is described.
\subsection{Printed text image generation}
    External information that researchers have defined in advance to describe languages is yet to be made available. Nevertheless, our main idea stems from that each language consists of character sets, which is well-known prior knowledge, no matter whether the scene text images of a language is seen or not. Hence, we leverage the character sets of each language to generate ``free-of-charge'' images of printed texts as the external information for text detection in our proposed problem setting.

    The procedure of printed text image generation is as follows:
    First, a text line is generated by randomly selecting several characters from the character set. Then, 15 text lines are synthesized to form a printed text image corresponding to a specific language. The font size of each text line is  randomly selected from 15pt to 40pt. Also, To make sure our method is insensitive to synthesized content, we conduct experiments using different content sources such as Wikipedia, Lyrics, and Bible texts to form the printed images. The results show that the different content sources do not significantly affect the text detection performance. Owing to many Chinese characters, we chose the 3000 most commonly used characters as the character set. The Japanese character set includes the syllabaries, hiragana, katakana, and kanji. However, we increase the probability of drawing syllabary samples to distinguish themselves from the Chinese characters. Moreover, we find the number of English character sets is small. We, therefore, selected 4000 commonly-used vocabularies as the sample set and randomly drew words to form printed text images to highlight the difference between English and other Latin languages. The procedure above allows free-of-charge printed text images to be synthesized efficiently.
    
\subsection{Synthetic text generation}\label{sec:text_gen}
    MLT19 \cite{icdar19} is used as our training dataset, containing ten languages and at least 1000 images in each language. Typically, each image in the dataset includes English texts and other-language texts. Such a distribution of language instances would have the detector favors English prediction. Therefore, we generate other synthetic instances that help balance the distribution of languages in scene images. First, we extracted several articles from BBC News in various countries and used the vocabulary in the articles as the words to synthesize texts for each language. Later, we ensure that each scene image in MLT19 contains up to a dozen synthetic words. The language of each synthetic word is randomly selected from the training language set. The color and font size are also randomly chosen, but synthetic words are not allowed to overlap with the original scene texts. It is worth noting that a part of the synthetic texts must have the same language as the real one in the current scene. It would help to learn the desired matching relationship for text detection. More discussions will be in the ablation study.
    
    To prevent the model from overfitting to the seen languages, we also generated printed text images and synthetic text instances in Thai, Gujarati, Amharic, and Hebrew as pseudo-unseen languages during the training phase only. These pseudo-unseen languages help to generalize our model and would not be used in the inference phase.
\subsection{Dynamic Guide}
    The Dynamic Guide (DG) aims to learn to generate representational attributes for different languages from printed text images. We take five printed text images, $X^{print}={\{x_{i}}\}_{i=1}^{5}\in{\mathbb{R}}^{5\times1\times{H}\times{W}}$ as input, and extract their features through a backbone (i.e., ResNet34\cite{resnet}) followed by an average-pooling, which is denoted as below.
    \begin{equation}
    \begin{split}
      B^{print}=avg(Backbone^{print}(X^{print}))\in{\mathbb{R}}^{5\times{C}\times{1}\times{1}}
      \label{eq:important}
    \end{split}
    \end{equation}
    To aggregate the 5 feature vectors (i.e.,$B^{print}$) to form a compact but rich representation for distinguishing different languages, we take the minimum, maximum, and average values of the five feature vectors, and then concatenate them to generate the language-specific representation $Match^{print}\in{\mathbb{R}}^{3C\times{1}\times{1}}$. Note that we also treat $Match^{print}$ as the learnable attributes (i.e., from the viewpoint of attribute-based zero-shot learning), which play important roles in seen/unseen text matching. Eventually, $Match^{print}$ is defined as 
    \begin{equation}
    \begin{split}
    Match^{print}=min(B^{print})\oplus{max(B^{print})}\oplus{avg(B^{print})}
        \label{eq:important}
    \end{split}
    \end{equation}
   To improve the ability to deal with unseen languages and further help the Dynamic Guide learn representative language features, we add an auxiliary branch to realize a language classifier and classify $Match_{print}$ into $c$ classes, where $c$ is the number of seen languages. For classification, $Match_{print}$ is passed through a tiny neural-net with two fully connected layers. Thereby, the language representation feature,  $Match_{print}$, would also be supervised by the binary classification loss. Since we expect our text detector to be applied to unseen languages, we avoid the extracted language feature to represent only the close-set languages used for training. To enhance language feature generalization, we thus generate printed text images of pseudo-unseen language as the other inputs for training. Without adding extra classes (e.g., unseen classes), we ask the probability outputs of $c$ seen languages to be as small as possible, which implicitly strengthens the ability to represent other open-set languages.
\subsection{Text Finder}
    The main purpose of the Text Finder (TF) is to obtain the differentiable features from scene images so that the text regions of the target language can be identified and the cluttered background can be removed in the following detection module. We divide TF into two subbranches; one is the language-agnostic text detection sub-branch, $Sub_{detect}^{scene}$, and the other is the language classifier, $Sub_{language}^{scene}$.

    The target of $Sub_{detect}^{scene}$ is to detect all text regions, excluding only the background, whether the language is seen or unseen. To achieve this goal, the model should learn the common text characteristics within various languages rather than extracting the specific text characteristics of the seen languages for training. Kim et al. \cite{objectness} proposed a classification-free object localization network (OLN) that estimates objectness by locating objects using a class-independent loss to address the problem of always ignoring new objects in the open world when detecting objects. Inspired by this work, we pre-train and later fixed the language-independent $Sub_{detect}^{scene}$ (i.e. including ResNet50\cite{resnet}, FPN, and the following ``Convs" block shown in Fig. \ref{fig:framework}) to learn generalized text features and identify the text regions without language classification. For $Sub_{detect}^{scene}$ pre-training, scene text images are inputted to the ResNet backbone (i.e. ResNet50\cite{resnet}) for feature extraction, followed by Feature Pyramid Network (FPN)\cite{fpn} for low- and high-level feature aggregation. The ``Convs" block implements convolutional neural networks to produce dense and pixel-wise text/non-text classification; it also localizes text by estimating centerness \cite{fcos}. Later, the dice loss for pixel-wise text classification and $L1$ loss for centerness estimation are used to supervise the $Sub_{detect}^{scene}$ pre-training. 

    The subbranch $Sub_{language}^{scene}$ is used to extract features of the text related to its language class. Its design concept is the same as the language classification of Dynamic Guide. The only difference is that $Sub_{language}^{scene}$ classifies the language pixel-wisely. When training $Sub_{language}^{scene}$, we generate the synthetic texts by the processing flow mentioned in Sec.~\ref{sec:text_gen} to augment the scene text images. Next, the synthetic scene text image passes through the fixed and shared ResNet backbone to extract multi-level features. After a series of convolutions, as illustrated in Fig. \ref{fig:framework}, the features from different levels are resized and concatenated to generate the language-specific feature maps $F_{mid}$ for pixel-wise language classification. When training $Sub_{language}^{scene}$, we also apply the aforementioned pseudo-unseen languages to avoid the preference of detecting only the seen language. For text regions belonging to pseudo-unseen languages, the $Sub_{language}^{scene}$ is forced to assign a small probability to the $c$ seen languages.

    Moreover, since $Sub_{detect}^{scene}$ is designed to catch all the text regions (i.e., including seen and unseen languages), we treat the output as a text mask, denoted as $Mask_{text}$. By multiplying the middle feature maps $F_{mid}$ and $Mask_{text}$ to generate $Match^{scene}$, the input of the following Language Mapper, we can significantly reduce false text detection caused by the cluttered background.
\subsection{Language Mapper}
    In the Language Mapper (LM), we have $Match^{print}$ from DG and $Match^{scene}$ from TF. The goal of the LM is to learn the corresponding relationship between the learnable printed text attributes and the scene text feature, which can also be said to take side information as a reference to check if there is a similarity between the scene text and its mentor.
    
    Since side information is replaceable and its corresponding printed text varies dynamically, well-used static convolution networks, whose parameters would be stuck to fit the seen language after training, become inappropriate. In order to meet the dynamically-changed input $X^{print}$, we adopt the input-dependent dynamic kernels as the medium of side information to search for the corresponding scene text feature, as shown in Fig. \ref{fig:framework} (i.e., language mapper). Remarkably, we design multiple Fully Connected (FC) networks to translate the language-specific attributes ($Match^{print}$) to corresponding kernel weights for multi-level dynamic convolutional networks. The process is defined as follows:
    \begin{equation}
    \begin{split}
        F_{kernel}=FC_{shared}(Match^{printed}), \\ 
        Kernel_{i}=FC_{ker_{i}}(F_{kernel}), i=1\sim3.
        \label{eq:important}
    \end{split}
    \end{equation}
The three kernels $Kernel_{i},i=1,2,3$ shown above gradually figure out the target text features in $Match^{scene}$; later, a static convolutional network is utilized to generate the outputs stored in two-channel maps. The first output channel indicates whether a pixel belongs to the same language as $X_{print}$. The second channel is a fine-grained estimation of the text regions, which are further refined based on the first channel. This helps to separate the close text instances. In experiments, we found that Japanese, Chinese, and Korean may sometimes be regarded as the same language. To distinguish their difference, we use pixel-wise contrastive learning, which aggregates and separates features according to target (positive) pixels and non-target(negative) pixels. Note that target and non-target pixels used for training can be determined via supervised labels. Specifically, after each dynamic convolution network, we apply the contrastive loss on the projected feature space. In this way, we can distinguish texts with similar appearances but from different languages.

As shown in Fig.~\ref{fig:framework}, our framework outputs two segmentation maps, denoted as kernel and text maps; they
are then post-processed to generate the final text bounding boxes. Note that the kernel map is a shrinking version of the
text field. By referring to the predicted text area (maps), we use the progressive scale expansion algorithm~\cite{pse} to expand the predicted kernel area. Finally, we compute the minimum area quadrilateral and extract the bounding boxes.

\subsection{Optimization}
    The whole network training includes two steps. In the first step, we pre-train the subnetwork $Sub_{detect}^{scene}$, defined in the TF module, to learn language-independent text features via the following objective function.
    \begin{equation}
        \mathcal{L}_{detect}=\lambda_{mask}\mathcal{L}_{mask}+\lambda_{center}\mathcal{L}_{center}.
        \label{eq:important}
    \end{equation}
     $\mathcal{L}_{mask}$ represents a dice loss for the prediction of text or non-text. $\mathcal{L}_{center}$ is $L1$ loss used to supervise whether a pixel is the center of a word. Here, we follow \cite{objectness} to determine the training label for text centerness. Also, we set both $\lambda_{mask}$ and $\lambda_{center}$ to 1 in this pre-training step.
    
     In the second step, we fixed $Sub_{detect}^{scene}$ and train the other parts based on ${L}_{match}$ defined as 
     \begin{equation}
     \begin{split}
        \mathcal{L}_{match}=\lambda_{scene}\mathcal{L}_{scene}+\lambda_{printed}\mathcal{L}_{printed}\\ +\lambda_{text}\mathcal{L}_{text}+\lambda_{kernel}\mathcal{L}_{kernel}+\Sigma_{i=1}^{3}\lambda_{con^{i}}\mathcal{L}_{con^{i}}.
    \label{eq:important}
    \end{split}
    \end{equation}
    Here, $\mathcal{L}_{scene}$ and $\mathcal{L}_{printed}$ are binary classification losses used to supervise the language type classifiers. However, $\mathcal{L}_{scene}$ is a pixel-wise classification loss used in the TF module. In contrast, $\mathcal{L}_{printed}$ is an image-wise loss appended in the DG module. On the other hand, we introduce the two pixel-wise dice losses, $\mathcal{L}_{text}$ and $\mathcal{L}_{kernel}$, at the output of the LM module to supervise the estimated text and shrunk text regions. Lastly, $\mathcal{L}_{con^{i}}$ is the contrastive loss also used in the LM module. In this step, we set $\lambda_{scene}$, $\lambda_{printed}$, $\lambda_{text}$, $\lambda_{kernel}$ and $\lambda_{(con^{i})},i=1,2,3$ as 0.01, 0.01, 1, 1, 0.05 before the first 100 epochs; later, we change $\lambda_{scene}$, $\lambda_{printed}$ and $\lambda_{(con^{i})}$ to 0.001, 0.001 and 0.005.

%% file: 4_experiment.tex
\section{Experiments}
\subsection{Datasets}
\noindent\textbf{ICDAR 2017 MLT (MLT17)}\cite{icdar17} comprises text embedded in natural scene images such as road signs and signboards. It contains 18,000 images, including 7200 training images, 1800 validation images, and 9000 testing images. The languages involved include Arabic, Bangla, Chinese, English, French, German, Italian, Japanese, and Korean, which might appear simultaneously in the same image. Besides, the dataset collected 2,000 images for each of the languages. 

\noindent\textbf{ICDAR 2019 MLT (MLT19)}\cite{icdar19} was built upon MLT17 with an additional language, Hindi. For better training, it consists of 20,000 real scene images and 277,000 synthetic ones (SynthTextMLT\cite{E2E}). Ten languages are involved in both real and synthetic images. We evaluated our method by the MLT17 validation set, which does not contain Hindi. Accordingly, we used MLT19 without Hindi as the training dataset to train our model.

\noindent\textbf{IIIT-ILST}\cite{malayalam} is a dataset and benchmark for scene text recognition for three Indian scripts - Devanagari, Telugu, and Malayalam. It comprises nearly 1000 images in the wild, which is suitable for scene text detection and recognition tasks. We take the Malayalam part as our unseen language for evaluation.

\begin{table*}
\centering
\caption{Comparison with the related works on the MLT17 validation set and Malayalam in IIIT-ILST. `*' means the unseen language of the method. `x' means that the method cannot detect unseen languages. We try three settings that use (a) Korean, (b) Chinese, and (c) Arabic as unseen languages alternatively.}
\begin{tabular}{c|ccccccc}
\hline
Method         & English & Arabic & Bangla & Chinese & Japanese & Korean   & Malayalam \\ \hline
E2E-MLT\cite{E2E}        & 55.43   & 55.431 & 3.027  & 50.594  & 12.9     & 32.715   & x         \\ \hline
MultiplexedOCR\cite{mul} & 83.284  & 80.074 & 78.104 & 56.251  & 70.986   & 67.862   & 11.932*   \\ \hline
(a)   Ours         & 84.031  & 80.952  & 81.76  & 83.585  & 76.896   & 72.479 * & 65.837*   \\
(b)   Ours         & 82.527  & 82.092  & 80.909 & 69.046* & 83.51    & 84.013   & 55.895*   \\
(c)   Ours         & 81.778  & 54.646* & 82.134 & 82.554  & 80.737   & 83.143   & 43.992*   \\ \hline
\end{tabular}
\label{tab:real}
\vspace{-1em}
\end{table*}

\begin{table*}[]
\centering
\caption{Comparison with related works on the synthetic MLT17 validation set and synthetic Malayalam in IIIT-ILST. The synthetic Malayalam dataset also includes all languages in the MLT17 dataset.}
\begin{tabular}{c|ccccccc}
\hline
Method         & English & Arabic & Bangla & Chinese & Japanese & Korean   & Malayalam \\ \hline
E2E-MLT\cite{E2E}        & 50.19   & 54.67  & 4.027  & 55.138  & 22.346   & 34.317   & x         \\ \hline
MultiplexedOCR\cite{mul} & 67.392  & 75.511 & 78.874 & 54.914  & 62.024   & 71.21    & 11.159*   \\ \hline
(a) Ours    & 71.76   & 74.949 & 79.044 & 74.351  & 71.556   & 50.534 * & 55.607*   \\
(b) Ours    & 63.119  & 71.811 & 80.393 & 48.905* & 63.472   & 75.385   & 42.497*   \\
(c) Ours    & 63.424  & 46.061* & 80.328 & 80.328  & 59.254   & 75.385   & 36.586*          \\ \hline
\end{tabular}
\label{tab:real&synth}
\vspace{-1em}
\end{table*}

\subsection{Implement detail and evaluation metrics}
    Since scene text images are more complex than printed ones, we choose ResNet-34 for the Dynamic Guide (DG) and deeper ResNet-50 for the Text Finder (TF), respectively. We train our model on the MLT19 training dataset, including 1000 images per language, and evaluate our model on the MLT17 validation set. Because the MLT17 dataset does not contain Hindi, we exclude it from our training and testing datasets. Precisely, the multiple languages we used for training or evaluation include Chinese, English, Arabic, Bangla, Japanese, and Korean. To ensure that we have enough training languages to learn the meta-mapping, which generalized the mapping relationship between the printed texts and language-specific representation, five of the six languages are chosen as the training set, and the rest is regarded as the unseen testing language. We also apply the k-fold cross-validation strategy for performance evaluation.

    We train our model in two stages. In the first stage, we train the language-agnostic text detection part of the scene branch on a 5K MLT19 dataset without any synthetic data. In the second stage, we train the rest of the network with the same 5K MLT19 training images but with additionally pasted synthetic text instances on these images. For data augmentation, we randomly crop the scene text images for all scene images in the datasets and resize them to ${832}\times{832}$. For the printed text images, we resize them to ${512}\times{512}$. The initial learning rate is set to $10^{-3}$ and divided by ten after 7,500 iterations.  Table \ref{tab:cost} presents our model's size, computational cost, and running speed.

    Note that our work focuses on pixel-wise text region segmentation instead of using regression-based methods to generate text bounding boxes directly. Thus, We converted the bounding boxes defined in ground truth as segmentation label maps by setting the pixels inside the bounding boxes to 1 and the others to 0 for a fair evaluation. In the following experiments, we also convert the detection results from the other compared methods into segmentation results. We evaluate our approach in two ways to demonstrate that we can detect the target language (i.e., specified by the painted images) in a testing scene image, including seen and unseen languages. First, we use the original MLT17 validation set as the test set. Furthermore, we manually paste synthetic texts on the testing scene images from the MLT17 validation set to construct a more challenging testing set, named ``synthetic MLT17". The language of these synthetic text instances includes not only the five seen languages and the unseen one but also two additional languages, Russian and Greek, which have not been seen. By doing so, we increase the difficulty of the new detection task.

\begin{table}[]
\caption{{\textbf{Computational cost and running speed.} The following table lists the number of trainable parameters and FLOPs of our MENTOR model.}}
\centering
\begin{tabular}{c|ccc}
\hline
 & Param (M)  & FLOPs (G) & FPS  \\ \hline
Text Finder  & 27.38 & 373.92  &    \\  \hline
Dynamic Guide   & 21.55 & 37.72 &   \\  \hline
Language Mapper   & 53.85 & 621.44 &  \\  \hline
MENTOR & 102.78 & 1033.08 & 3.25   \\  \hline
\end{tabular}
\label{tab:cost}
\vspace{-1em}
\end{table}

\subsection{Comparions}
    Our  comparison is with the supervised methods E2E-MLT\cite{E2E} and MultiplexerOCR\cite{mul}, which include both text detection and language identification. E2E-MLT was the first published multi-language text spotting method, trained on public scene text datasets and the synthetic multi-language dataset ``SynthTextMLT''. With 245,000 images in 9 languages, text detection and language recognition were trained. Note that E2E-MLT employs a majority voting scheme to perform language identification based on each character's language output
    MultiplexerOCR, the SOTA scene text spotting method, is trained on the MLT17 train set, the MLT19 train set, and SynthTextMLT, using over 273K images. Language recognition is accomplished by extracting the masked pooled feature from their proposed detection module. 
    
    In Table \ref{tab:real}, we compare our methods with the two SOTA methods on the MLT17 validation set and show the language-wise F-score results. Both our methods (a) and (b), which use Korean and Chinese separately as the unseen language, outperform the two supervised methods. In setting (c), where Arabic is an ``unseen'' language, we achieve comparable performance to E2E-MLT, which uses Arabic as a ``seen'' language. Additionally, our results demonstrate superior performance for the seen languages, particularly Chinese, Japanese, and Korean.
    
    We also evaluated Malayalam from the IIIT-ILST dataset as an unseen language in Table \ref{tab:real}. As MultiplexerOCR does not have a category for unseen languages, we considered any text with a detection confidence lower than 0.5 as an unseen language category. 
    In this case, MultiplexerOCR only achieved an 11.159\% detection rate for unseen languages, and even with adjustment of the confidence threshold, the optimal F-score only increased by about 2\%. By contrast, our Malayalam results demonstrate greater effectiveness in handling unseen languages than other methods.
    
    Furthermore, we provide evaluation results for the synthetic MLT17 validation set in Table \ref{tab:real&synth}. This test scenario is even more challenging, but our results show that our performance remains competitive with other supervised learning methods even when Chinese is designated as an unseen language. On the other hand, Korean is a language that can be easily confused with Chinese and Japanese, yet we still outperformed E2E-MLT when Korean was designated as an unseen language. Arabic text has a unique line-like appearance and can be more similar to a background texture than actual text, making it challenging to treat Arabic as an unseen language.

\begin{table}[]
\caption{{\textbf{Number of printed text images.}}}
\centering
\begin{tabular}{c|ccccc}
\hline
& 1 & 5 & 10 & 15 & 20 \\ \hline
English & 63.202 & 63.276 & 63.38 & 62.989 & 63.032 \\ \hline
Korean & 77.338 & 77.379 & 77.537 & 77.331 & 77.518 \\ \hline
Chinese (unseen) & 48.906 & 48.939 & 48.939 & 48.939 & 48.907\\ \hline
\end{tabular}
\label{tab:printext}
\vspace{-1em}
\end{table}

\subsection{Ablation study}
    \textbf{Number of printed text images \& number of progressive comparison.}
    We have experimented with different numbers of printed text images as the input of the DG module, where the results for unseen text detection summarized in Table \ref{tab:printext} demonstrate that our model is insensitive to different settings. Regarding the number of progressive comparisons (i.e., the number of dynamic kernels), we once set it to 1 at the beginning of our model development; however, it did not function well enough. Afterward, we experimented with increasing the number of dynamic kernels for progressive comparison. Considering the training parameters required by the dynamic kernel (cf. Table in \cite{pse}) and the overall training time, we finally use 3 dynamic matching layers.
    
    \textbf{Language-agnostic text detection.} 
    In Text Finder, the output of language-independent text detection, $Sub_{detect}^{scene}$, is used as a mask to filter complex backgrounds. We hope the detector can identify as many text regions as possible but does not need the ability to distinguish languages. Thus, its function is to learn the common characteristics among all possible languages. We use real scene text images as training data to train $Sub_{detect}^{scene}$. However, to verify its ability to detect unseen languages, we tested $Sub_{detect}^{scene}$ with the synthetic MLT17 validation set, including text instances from the five seen languages and three unseen languages explained above. The results in Table \ref{tab:detect} show the success rates of text detection for each specific language. 
    According to the results, the performance of detecting texts from unseen languages is comparable to that of seen languages, meaning that the proposed method realizes the goal of language-independent text detection. 
    
\begin{figure}[t]
  \centering
  \includegraphics[width=1\linewidth]{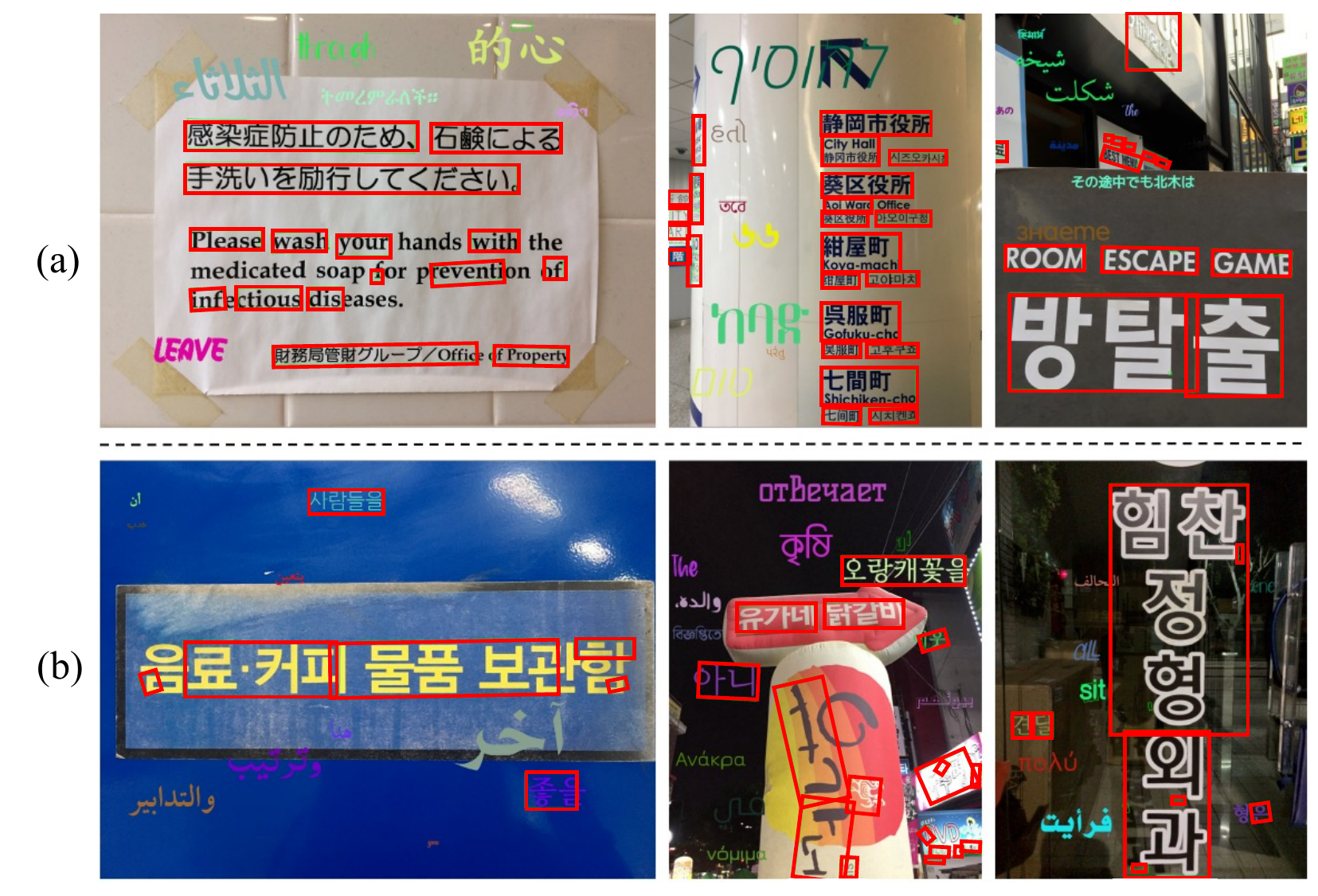}
  \caption{(a) Detection results trained with disjoint synthetic scene data. The target language from left to right is Japanese, Japanese, and Korean. All the real scene text is detected regardless of its language, which reveals that a good matching relationship is not well-learned due to the insufficient training of the language mapper. (b) The model was trained on intersection synthetic data. Korean is the unseen language. The detector can detect both synthesized and real scene texts in Korean without being confused with other languages.}
  \label{fig:compare} 
  \vspace{-1em}
\end{figure}

\begin{table}[]
\caption{{\textbf{Language-agnostic detection on synthetic MLT17 validation set.} The language-agnostic detector
is trained with real-scene images containing five seen languages. Russian and Greek texts are synthetic.}}
\centering
\begin{tabular}{c|c|c}
\hline
Language & seen/unseen & Percentage (\%) \\ \hline
English  & seen        & 86.378          \\
Arabic   & seen        & 79.667          \\
Bangla   & seen        & 80.511          \\
Japanese & seen        & 86.565          \\
Chinese  & seen        & 88.294          \\
Korean   & unseen      & 80.334          \\
Russian  & unseen      & 78.517          \\
Greek    & unseen      & 75.155          \\ \hline
\end{tabular}
\label{tab:detect}
\vspace{-2em}
\end{table}
   
    \textbf{Intersection synthetic data vs. disjoint synthetic data.}
    As mentioned in Sec. 3.2, we synthesize text for the second-step training to make MENTOR more general to the unseen languages. During synthesizing texts, we find that parts of the generated texts should come from the same language as the ones in the real scene. For analysis, we compare two synthetic text sets while training: the ``intersection dataset'' and the ``disjoint dataset''. For example, if Chinese and English phrases appear in a scene image and we generate synthetic texts from the two languages, it becomes ``intersection data'';  otherwise, we call it ``disjoint data". To prevent ``MENTOR'' from detecting only real scene texts and ignoring synthetic texts, we find the intersection dataset can help our model learn the intrinsic properties of texts instead of learning the style difference between the real and synthetic texts. Contrastively, if using `disjoint datasets'' for training, the TF module only detects the real scene texts, the DG module cannot generate the language-specific representation, and the LM fails to detect texts by matching.
    
    Table \ref{tab:balance}. shows the F-score of training with intersection data and disjoint data. When using disjoint data, we find an unreasonable result that the performance of the unseen Korean is better than other seen languages. By checking the detection results shown in Fig.~\ref{fig:compare}(a), we find the model is falsely trained to detect only the real scene texts rather than learning the intrinsic property of languages. After training with intersection data, we get more reasonable performance. Fig.~\ref{fig:compare}(b) is an example.

\begin{table}[]
\caption{\textbf{Quantitative detection results under intersection synthetic data vs. disjoint synthetic data.} Using disjoint synthetic data, our model learns to detect scene texts according to text style. In the unseen Korean case, the Korean scene texts occupy more than the texts from other languages, leading to an illusion of good detection results. In contrast, the model can be well-trained for multilingual text detection based on intersection data.}
\scalebox{0.9}{
\begin{tabular}{c|cccccc}
\hline
\begin{tabular}[c]{@{}c@{}}Train\\ data\end{tabular} & Japanese & English & Arabic & Chinese & Bangla & \begin{tabular}[c]{@{}c@{}}Korean\\ (unseen)\end{tabular} \\ \hline
Disjoint                                            & 53.99    & 58.07   & 57.39  & 71.58   & 71.15  & 69.4                                                      \\ \cline{1-1}
Intersection                                              & 71.56    & 71.76   & 74.94  & 74.35   & 79.04  & 50.53                                                     \\ \hline
\end{tabular}
}
\label{tab:balance}
\vspace{-1.5em}
\end{table}

%% file: 5_conclusion.tex
\section{Conclusion}
We introduced a new problem setting for multilingual scene text detection and proposed a novel method for detecting both seen and unseen languages. To identify the text regions of an unseen language without model re-training and labeled data, our ``MENTOR'' introduces ``DG'', a dynamic and learnable module, to extract language-specific features for each language. Moreover, the ``TF'' module detects seen and unseen text regions and extracts pixel-wise language features from the input image. Finally, by comparing the language-specific features (from DG) and pixel-wise scene text features (from TF), we can identify the text regions of unseen languages via our ``LM'' module. The experiments demonstrated the effectiveness of our ``MENTOR'' network.

%% file: root.bbl
\begin{thebibliography}{10}\itemsep=-1pt

\bibitem{craft}
Youngmin Baek, Bado Lee, Dongyoon Han, Sangdoo Yun, and Hwalsuk Lee.
\newblock Character region awareness for text detection.
\newblock In {\em IEEE Conference on Computer Vision and Pattern Recognition
  (CVPR)}, 2019.

\bibitem{E2E}
Michal Bušta, Yash Patel, and Jiri Matas.
\newblock E2e-mlt - an unconstrained end-to-end method for multi-language scene
  text.
\newblock {\em ArXiv:1801.09919}, 2018.

\bibitem{position}
Peirui Cheng, Yuanqiang Cai, and Weiqiang Wang.
\newblock A direct regression scene text detector with position-sensitive
  segmentation.
\newblock {\em IEEE Transactions on Circuits and Systems for Video Technology},
  2020.

\bibitem{progress}
Pengwen Dai, Sanyi Zhang, Hua Zhang, and Xiaochun Cao.
\newblock Progressive contour regression for arbitrary-shape scene text
  detection.
\newblock In {\em IEEE Conference on Computer Vision and Pattern Recognition
  (CVPR)}, 2021.

\bibitem{resnet}
Kaiming He, Xiangyu Zhang, Shaoqing Ren, and Jian Sun.
\newblock Deep residual learning for image recognition.
\newblock {\em ArXiv:1512.03385}, 2015.

\bibitem{most}
Minghang He, Minghui Liao, Zhibo Yang, Humen Zhong, Jun Tang, Wenqing Cheng,
  Cong Yao, Yongpan Wang, and Xiang Bai.
\newblock Most: A multi-oriented scene text detector with localization
  refinement.
\newblock In {\em IEEE Conference on Computer Vision and Pattern Recognition
  (CVPR)}, 2021.

\bibitem{mul}
Jing Huang, Guan Pang, Rama Kovvuri, Mandy Toh, Kevin~J Liang, Praveen
  Krishnan, Xi Yin, and Tal Hassner.
\newblock A multiplexed network for end-to-end, multilingual ocr.
\newblock {\em ArXiv:2103.15992}, 2021.

\bibitem{rcnn}
Zhida Huang, Zhuoyao Zhong, Lei Sun, and Qiang Huo.
\newblock Mask r-cnn with pyramid attention network for scene text detection.
\newblock In {\em IEEE Winter Conference on Applications of Computer Vision
  (WACV)}, 2019.

\bibitem{objectness}
Dahun Kim, Tsung-Yi Lin, Anelia Angelova, In~So Kweon, and Weicheng Kuo.
\newblock Learning open-world object proposals without learning to classify.
\newblock {\em IEEE Robotics and Automation Letters (RAL)}, 2022.

\bibitem{sia}
Gregory Koch, Richard Zemel, and Ruslan Salakhutdinov.
\newblock Siamese neural networks for one-shot image recognition.
\newblock In {\em International Conference on Machine Learning (ICML)
  Workshops}, 2015.

\bibitem{iap}
Christoph~H. Lampert, Hannes Nickisch, and Stefan Harmeling.
\newblock Attribute-based classification for zero-shot visual object
  categorization.
\newblock {\em IEEE Transactions on Pattern Analysis and Machine Intelligence
  (TPAMI)}, 2014.

\bibitem{mask}
Minghui Liao, Guan Pang, Jing Huang, Tal Hassner, and Xiang Bai.
\newblock Mask textspotter v3: Segmentation proposal network for robust scene
  text spotting.
\newblock In {\em European Conference on Computer Vision (ECCV)}, 2020.

\bibitem{fpn}
Tsung-Yi Lin, Piotr Dollár, Ross Girshick, Kaiming He, Bharath Hariharan, and
  Serge Belongie.
\newblock Feature pyramid networks for object detection.
\newblock {\em ArXiv:1612.03144}, 2016.

\bibitem{fots}
Xuebo Liu, Ding Liang, Shi Yan, Dagui Chen, Yu Qiao, and Junjie Yan.
\newblock Fots: Fast oriented text spotting with a unified network.
\newblock {\em ArXiv:1801.01671}, 2018.

\bibitem{context}
Xi Liu, Gaojing Zhou, Rui Zhang, and Xiaolin Wei.
\newblock An accurate segmentation-based scene text detector with context
  attention and repulsive text border.
\newblock In {\em IEEE/CVF Conference on Computer Vision and Pattern
  Recognition (CVPR) Workshops}, 2020.

\bibitem{bezier}
Yuliang* Liu, Hao* Chen, Chunhua Shen, Tong He, Lianwen Jin, and Liangwei Wang.
\newblock Abcnet: Real-time scene text spotting with adaptive bezier-curve
  network.
\newblock {\em ArXiv:2002.10200}, 2020.

\bibitem{malayalam}
M. {Mathew}, M. {Jain}, and C.~V. {Jawahar}.
\newblock Benchmarking scene text recognition in devanagari, telugu and
  malayalam.
\newblock In {\em IAPR International Conference on Document Analysis and
  Recognition (ICDAR) Workshops}, 2017.

\bibitem{icdar19}
Nibal Nayef, Yash Patel, Michal Busta, Pinaki~Nath Chowdhury, Dimosthenis
  Karatzas, Wafa Khlif, Jiri Matas, Umapada Pal, Jean-Christophe Burie,
  Cheng-lin Liu, and Jean-Marc Ogier.
\newblock Icdar2019 robust reading challenge on multi-lingual scene text
  detection and script identification - rrc-mlt.
\newblock In {\em IAPR International Conference on Document Analysis and
  Recognition (ICDAR)}, 2019.

\bibitem{icdar17}
Nibal Nayef, Fei Yin, Imen Bizid, Hyunsoo Choi, Yuan Feng, Dimosthenis
  Karatzas, Zhenbo Luo, Umapada Pal, Christophe Rigaud, Joseph Chazalon, Wafa
  Khlif, Muhammad~Muzzamil Luqman, Jean-Christophe Burie, Cheng-lin Liu, and
  Jean-Marc Ogier.
\newblock Icdar2017 robust reading challenge on multi-lingual scene text
  detection and script identification - rrc-mlt.
\newblock In {\em IAPR International Conference on Document Analysis and
  Recognition (ICDAR)}, 2017.

\bibitem{proto}
Jake Snell, Kevin Swersky, and Richard~S. Zemel.
\newblock Prototypical networks for few-shot learning.
\newblock {\em ArXiv:1703.05175}, 2017.

\bibitem{ltc}
Flood Sung, Yongxin Yang, Li Zhang, Tao Xiang, Philip H.~S. Torr, and
  Timothy~M. Hospedales.
\newblock Learning to compare: Relation network for few-shot learning.
\newblock {\em ArXiv:1711.06025}, 2017.

\bibitem{fcos}
Zhi Tian, Chunhua Shen, Hao Chen, and Tong He.
\newblock Fcos: Fully convolutional one-stage object detection.
\newblock {\em ArXiv:1904.01355}, 2019.

\bibitem{vatti}
Bala~R. Vatti.
\newblock A generic solution to polygon clipping.
\newblock {\em Communications of the ACM}, 1992.

\bibitem{relation}
Oriol Vinyals, Charles Blundell, Timothy Lillicrap, Koray Kavukcuoglu, and Daan
  Wierstra.
\newblock Matching networks for one shot learning.
\newblock {\em ArXiv:1606.04080}, 2016.

\bibitem{pse}
Wenhai Wang, Enze Xie, Xiang Li, Wenbo Hou, Tong Lu, Gang Yu, and Shuai Shao.
\newblock Shape robust text detection with progressive scale expansion network.
\newblock In {\em IEEE Conference on Computer Vision and Pattern Recognition
  (CVPR)}, 2019.

\bibitem{explore}
Shilian Wu, Wei Zhai, Yongrui Li, Kewei Wang, and Zengfu Wang.
\newblock On exploring and improving robustness of scene text detection models,
  2021.

\bibitem{accurate}
Shanyu Xiao, Liangrui Peng, Ruijie Yan, Keyu An, Gang Yao, and Jaesik Min.
\newblock Sequential deformation for accurate scene text detection.
\newblock In {\em European Conference on Computer Vision (ECCV)}, 2020.

\bibitem{pyramid}
Enze Xie, Yuhang Zang, Shuai Shao, Gang Yu, Cong Yao, and Guangyao Li.
\newblock Scene text detection with supervised pyramid context network.
\newblock {\em ArXiv:1811.08605}, 2018.

\bibitem{geometry}
Youjiang Xu, Jiaqi Duan, Zhanghui Kuang, Xiaoyu Yue, Hongbin Sun, Yue Guan, and
  Wayne Zhang.
\newblock Geometry normalization networks for accurate scene text detection.
\newblock In {\em IEEE International Conference on Computer Vision (ICCV)},
  2019.

\bibitem{look}
Chengquan Zhang, Borong Liang, Zuming Huang, Mengyi En, Junyu Han, Errui Ding,
  and Xinghao Ding.
\newblock Look more than once: An accurate detector for text of arbitrary
  shapes.
\newblock In {\em IEEE Conference on Computer Vision and Pattern Recognition
  (CVPR)}, 2019.

\bibitem{anchor}
Zhuoyao Zhong, Lei Sun, and Qiang Huo.
\newblock An anchor-free region proposal network for faster r-cnn based text
  detection approaches.
\newblock {\em ArXiv:1804.09003}, 2018.

\end{thebibliography}
